\documentclass{article}

\usepackage[preprint,nonatbib]{neurips_2019}

\usepackage[utf8]{inputenc} % allow utf-8 input
\usepackage[T1]{fontenc}    % use 8-bit T1 fonts
\usepackage{url}            % simple URL typesetting
\usepackage{booktabs}       % professional-quality tables
\usepackage{amsfonts}       % blackboard math symbols
\usepackage{nicefrac}       % compact symbols for 1/2, etc.
\usepackage{microtype}      % microtypography
\usepackage{algorithm,algorithmic,amsmath,amssymb,amsthm}
\usepackage{graphicx}
\usepackage{multicol}
\usepackage{multirow}
\usepackage{wrapfig}

\newcommand{\eat}[1]{}

\DeclareMathOperator*{\argmin}{argmin}
\newcommand{\bx}{\mathbf{x}}
\newcommand{\bb}{\mathbf{b}}
\newcommand{\by}{\mathbf{y}}
\newcommand{\bz}{\mathbf{z}}
\newcommand{\bu}{\mathbf{u}}

\newcommand{\bbR}{\mathbb{R}}
\newcommand{\bbE}{\mathbb{E}}

\newcommand{\cX}{\mathcal{X}}

\newcommand{\bdelta}{\boldsymbol{\delta}}

\title{Model Agnostic Contrastive Explanations for Structured Data}

\author{
 Amit Dhurandhar\thanks{First four authors have equal contribution.}\\
  IBM Research \\
  Yorktown Heights, NY 10598 \\
  \texttt{adhuran@us.ibm.com} \\
  \And
  Tejaswini Pedapati \\
IBM Research \\
  Yorktown Heights, NY 10598 \\
  \texttt{tejaswinip@us.ibm.com} \\
  \And
  Avinash Balakrishnan\\
  IBM Research \\
  Yorktown Heights, NY 10598 \\
  \texttt{avinash.bala@us.ibm.com} \\
  \And
  Pin-Yu Chen\\
  IBM Research \\
  Yorktown Heights, NY 10598 \\
  \texttt{pin-yu.chen@ibm.com} \\
  \And
  Karthikeyan Shanmugam\\
  IBM Research \\
  Yorktown Heights, NY 10598 \\
  \texttt{karthikeyan.shanmugam2@ibm.com} \\
  \And
  Ruchir Puri\\
  IBM Research \\
  Yorktown Heights, NY 10598 \\
  \texttt{ruchir@us.ibm.com} 
}

\begin{document}
% \nipsfinalcopy is no longer used

\maketitle

\begin{abstract}
Recently, a method \cite{CEM} was proposed to generate contrastive explanations for differentiable models such as deep neural networks, where one has complete access to the model. In this work, we propose a method, Model Agnostic Contrastive Explanations Method (MACEM), to generate contrastive explanations for \emph{any} classification model where one is able to \emph{only} query the class probabilities for a desired input. This allows us to generate contrastive explanations for not only neural networks, but models such as random forests, boosted trees and even arbitrary ensembles that are still amongst the state-of-the-art when learning on structured data \cite{kaggle}. Moreover, to obtain meaningful explanations we propose a principled approach to handle real and categorical features leading to novel formulations for computing pertinent positives and negatives that form the essence of a contrastive explanation. A detailed treatment of the different data types of this nature was not performed in the previous work, which assumed all features to be positive real valued with zero being indicative of the least interesting value. We part with this strong implicit assumption and generalize these methods so as to be applicable across a much wider range of problem settings. We quantitatively and qualitatively validate our approach over 5 public datasets covering diverse domains.
\end{abstract}
%\vspace{-0.2cm}
\section{Introduction}
Given the wide spread use of deep networks \cite{gan} across various applications and their black box nature, explainability in artificial intelligence (XAI) has been one of the problems at the forefront in AI research recently \cite{montavon2017methods,patternet,lime,CEM,simple}. Darpa's call for creating interpretable solutions \cite{xai} and the General Data Protection Regulation (GDPR) passed in Europe \cite{gdpr} which requires businesses to provide understandable justifications to their users for decisions that may affect them has made this need even more acute.

There have been many (posthoc) interpretability methods proposed to interpret decisions of neural networks \cite{Ormas,patternet,saliency,bach2015pixel,CEM,simple} which assume complete access to the model. Locally interpretable model-agnostic explanation method (LIME) \cite{lime} is amongst the few that can provide local explanations for any model with just query access. %In other words, LIME just needs to be able to query the classification model and based on its outputs can generate an explanation. 
This is an extremely attractive feature as it can be used in settings where the model owner may not want to expose the inner details of the model but may desire local explanations using say a remote service. Another application is in interpreting decisions not just of neural networks but other models such as random forests, boosted trees and ensembles of heterogeneous models which are known to perform quite well in many domains that use structured data \cite{kaggle}.

In this paper, we thus propose the model agnostic contrastive explanations method (MACEM) that requires only query access to the classification model with particular focus on structured data. Structured data can be composed of real and categorical features, and we provide a principled way of creating contrastive explanations for such data. Contrastive explanations are a rich form of explanation where one conveys not only what is (minimally) sufficient to justify the class of an input i.e. pertinent positives (PPs), but also what should be (minimally) necessarily absent to maintain the original classification i.e. pertinent negatives (PNs) \cite{CEM}. Such explanations are commonly used in social settings as well as in domains such as medicine and criminology \cite{pertneg}. For example, a patient with symptoms of cough, cold and fever (PPs) could have flu or pneumonia. However, the absence of chills or mucous (PNs) would indicate that the person has flu rather than pneumonia. Thus, in addition to the symptoms that were present, the symptoms that are absent are also critical in arriving at a decision. As such, these type of explanations are also sought after in the financial industry where in the recently completed FICO explainability challenge \cite{FICO} it was explicitly stated that if a loan was rejected it would be highly desirable for an explanation to elucidate what changes to the loan application (i.e. input) would have led to its acceptance. Not to mention the data in the challenge was in structured format.

Additionally, working with experts across multiple industries (finance, healthcare, manufacturing, utility) we have found that they want explanation methods to satisfy two main criteria: a) be model agnostic so that one can explain a model on their private cloud through just query access and b) be trustworthy in that the method closely captures what the model is trying to do. For b) they are moving away from proxy model approaches such as LIME, since it does NOT meet their regulatory standards. %This is highlighted by statements such as "... the model has to explain itself ..." by auditing agencies where using even the term proxy to explain something is interpreted as violating that statement. 
\begin{wrapfigure}{r}{0.55\textwidth}
\vspace{-0.5cm}
\begin{center}
    \includegraphics[width=0.55\textwidth]{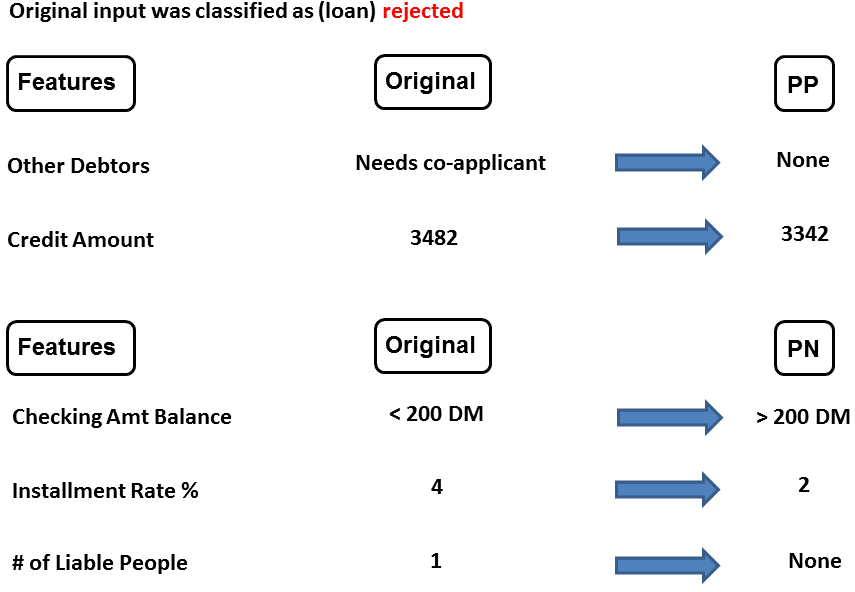}  
  \caption{Above we see an example explanation for a loan application from the German Credit dataset that was rejected by a black box model which was a tree. We depict the important features for PPs and PNs. Our PPs convey that even if the person didn't need a co-applicant and had lower credit card debt the application would still be rejected. In contrast, our PNs inform us that if the persons checking amount had more money, the loan installment rate was lower and there were no people that he/she was responsible for then the loan would have been accepted.}
  \label{VisualPPPN}
  \end{center}
\vspace{-0.5cm}
\end{wrapfigure}
Being contrastive is also extremely useful as it helps them understand sensitivities of the model decisions. Given this we believe our contribution is timely and significant.

In previous works \cite{CEM}, a method to produce such explanations was proposed. However, the method was restricted to differentiable models such as deep neural networks and strong (implicit) assumptions were made in terms of the semantics of the data used to train the models and obtain explanations. In particular, following are the key differences between our current and the prior work:\\
%\begin{itemize}
     \textbf{i) Gradients not available:} In this work we want to create contrastive explanations with only query access to any classification model. This is a significant step given that the prior work a) assumed complete access to the model and b) could be used only for differentiable models like deep neural networks.\\% Our method can be used for any classification model (viz. decision trees, forests, ensembles), where we estimate gradients (with theoretically bounded bias) using only oracle access.\\
     \textbf{ii) Using (and estimating) base values:} To compute PPs and PNs one needs to know what it means for a feature to be absent i.e., what value for a feature indicates there is no signal or is essentially the least interesting value for that feature. We refer to such values as \emph{base values}. In the prior work on contrastive explanations \cite{CEM} the value 0 for a feature was considered as the base value, which can be unrealistic. %However, this may not be the case for many features especially those that are categorical. 
     Ideally, the user should provide us with these values. In this paper we adapt our methods to utilize such base values and also propose ways to estimate them in situations that they are not provided. It is important to note here that existence base values is implicitly assumed in most of explainability research. For example, explanations for images involve highlighting/selecting pixels \cite{saliency,bach2015pixel,lime} which implicitly assumes that a blank image is zero information although it too may be classified in a class with high confidence.\\
     \textbf{iii) Handling categorical features:} In the prior work all features were considered to be real valued and no special consideration was given to handle categorical features. However, in this work we remain cognizant to the fact that categorical features are fundamentally different than real valued ones and propose a principled as well as scalable approach to handle them for our explanations.\\
     \textbf{iv) Computing PPs and PNs:} Given the above differences we propose new ways of computing PPs and PNs that are consistent with their intuitive definitions mentioned before. \emph{As such, we define a PP for an input $\bx$ as the sparsest example (w.r.t. base values) whose feature values are no farther from the base values than those of $\bx$, with it lying in the same class as $\bx$. Consequently, a PN for $\bx$ is defined as an example that is closest to $\bx$ but whose feature values are at least as far away from the base values as those of $\bx$ with it lying in a different class}. Important features for an example PP and PN for a loan application in the German Credit dataset are depicted in figure \ref{VisualPPPN}.
    
    %Although conceptually similar to the previous work, the details of the methods to compute these are quite different as can be witnessed in the later sections.
%\end{itemize}

%\begin{figure}[t]
 % \centering  
  %    \includegraphics[width=0.8\textwidth]{Visual_PP_PN.png}  
  %\caption{***** Replace this with PP/PN example from ppt for german credit **********
  %Above we see a visual depiction of PPs and PNs. Depending on where the input $\bx$ features lie relative to the base values the corresponding PP and PN features will be above or below the base value. However, for both PPs and PNs we are searching for the sparsest examples (w.r.t. base values) that are consistent with their interpretations (i.e. to find PNs we want to add (minimal) information to features of $\bx$ so that the predicted class changes, where addition is defined as moving away from the base values).}
  %\label{VisualPPPN}
%\vspace{-0.25cm}
%\end{figure}

\vspace{-0.3cm}
\section{Related Work}
\vspace{-0.3cm}
Trust and transparency of AI systems has received a lot of attention recently \cite{xai}. Explainability is considered to be one of the cornerstones for building trustworthy systems and has been of particular focus in the research community \cite{lipton2016mythos,tip}. Researchers are trying to build better performing interpretable models \cite{decl,twl,Caruana:2015,irt,bastani2017interpreting,simple} as well as improved methods to understand black box models such as deep neural networks \cite{lime,bach2015pixel,CEM}.

The survey \cite{montavon2017methods} which is mainly focused on deep learning explainability methods looks broadly at i) prototype selection methods \cite{nguyen2016synthesizing,nguyen2016multifaceted} to explain a particular class and ii) methods that highlight relevant features for a given input \cite{bach2015pixel,patternet,lime,unifiedPI}. There are other works that fall under (i) such as \cite{l2c,proto} as well as those that fall under (ii) for vision \cite{selvaraju2016grad,saliency,Ormas} and NLP applications \cite{lei2016rationalizing}. Most of these works though do not provide contrastive explanations in a model agnostic setting. There are also interesting works which try to quantify interpretability \cite{QEval,tip} and suggest methods for doing so.
 
Two of the most relevant recent works besides \cite{CEM} which we have already contrasted with are \cite{anchors,symbolic}. In \cite{anchors}, the authors try to find sufficient conditions to justify a classification that are global in nature. For example, the presence of the word "bad" in a sentence would automatically indicate negative sentiment irrespective of the other words. As such, they do not find input specific minimally sufficient values that would maintain the classification or minimal values that would change classification. Such global anchors also may not always be present in the data that one is interested in. The other work \cite{symbolic} tries to provide (stable) suggestions more than local explanations for decisions based on a neural network. Moreover, the approach is restricted to neural networks using rectified linear units and is feasible primarily for smallish to medium sized neural networks in asymmetric binary settings, where suggestions are sought for a specific class (viz. loan rejected) and not the other (viz. loan accepted).

\vspace{-0.3cm}
\section{MACEM Method}
\vspace{-0.3cm}
 Let $\cX$ denote the feasible data space and let $(\bx_0,t_0)$ denote an input example $\bx_0 \in \cX$ and its inferred class label $t_0$ obtained from a black-box classification model. The modified example $\bx \in \cX$ based on $\bx_0$ is defined as $\bx=\bx_0 + \bdelta$, where $\bdelta$ is a perturbation applied to $\bx_0$. Our method of finding pertinent positives/negatives is formulated as an optimization problem over the perturbation variable $\bdelta$ that is used to explain the model's prediction results. We denote the prediction of the model on the example $\bx$ by $\zeta(\bx)$, where $\zeta(\cdot)$ is any function that outputs a vector of confidence scores over all classes, such as the log value of prediction probability. Let $c,~\beta,~\gamma$ be non-negative regularization parameters.

\begin{algorithm}[htbp]
    \caption{Model Agnostic Contrastive Explanations Method (MACEM)}
    \label{cem}
\begin{algorithmic}
\STATE \textbf{Input:} Black box model $\mathcal{M}$, base values $\bb$, allowed space $\cX$, example $(\bx_0,t_0)$,  and (estimate of) input probability distribution $p(x)$ (optional).\\
%\STATE
\STATE 1) Find PPs $\bdelta^{\textnormal{pos}}$ by solving equation \ref{eqn_per_pos} and PNs $\bdelta^{\textnormal{neg}}$ by solving equation \ref{eqn_per_neg}.
%\STATE $\bdelta^{\textnormal{pos}}\gets\argmin_{\bdelta \in \Delta_{PP}}~c \cdot f^{\textnormal{pos}}_{\kappa}(\bx_0,\bdelta)+ \beta \|\bx_0 + \bdelta - \bb \|_1 + \|\bx_0 + \bdelta - \bb\|_2^2-\gamma p(\bx_0 + \bdelta).$
%\STATE 2) Find PNs $\bdelta^{\textnormal{neg}}$ by solving equation \ref{eqn_per_neg}.
%\STATE $\bdelta^{\textnormal{neg}}\gets\argmin_{\bdelta \in \Delta_{PN}}~c \cdot f^{\textnormal{neg}}_{\kappa}(\bx_0,\bdelta)+ \beta \|\bdelta  \|_1 + \|\bdelta \|_2^2 \nonumber - \gamma p(\bx_0 + \bdelta).$%\STATE

 \STATE 2) Return $\bdelta^{\textnormal{pos}}$ and $\bdelta^{\textnormal{neg}}$. \COMMENT{Explanation: The input $\bx_0$ would still be classified into class $t_0$ even if it were (closer to base values) $\bx_0+\bdelta^{\textnormal{pos}}$. However, its class would change if it were perturbed (away from original values) by $\bdelta^{\textnormal{neg}}$, i.e., if it became $\bx_0+\bdelta^{\textnormal{neg}}$. %Code provided in supplement.
}
\end{algorithmic}
\end{algorithm}
\vspace{-0.5cm}
\subsection{Computing Pertinent Positives}
\vspace{-0.2cm}
Assume an example $\bx_0$ has $d$ features each with base values $\{b_i\}_{i=1}^d$. Let $\Delta_{PP}$ denote the space $\{\bdelta:  | \bx_0+\bdelta - \bb | 	\preceq   | \bx_0 - \bb |  \textnormal{~and~} \bx_0 + \bdelta \in \cX  \}$, where $\bb=[b_1,\ldots,b_d]$, and $|\cdot|$ and $\preceq$ implies element-wise absolute value and inequality, respectively. To solve for PP, we propose the following:

%\vspace{-0.5cm}
\begin{align}
\label{eqn_per_pos}
\bdelta^{\textnormal{pos}} \leftarrow \argmin_{\bdelta \in \Delta_{PP}}~&c \cdot \max \{  \max_{i \neq t_0} [\zeta(\bx_0+\bdelta)]_i - [\zeta(\bx_0+\bdelta)]_{t_0}, - \kappa   \}+ \beta \|\bx_0 + \bdelta - \bb \|_1 \nonumber\\ &+ \|\bx_0 + \bdelta - \bb\|_2^2-\gamma p(\bx_0 + \bdelta).
\end{align}
 The first term %$f^{\textnormal{pos}}_{\kappa}(\bx_0,\bdelta)$
 is a designed loss function that encourages the modified example $\bx=\bx_0+\bdelta$ relative to the base value vector $\bb$, defined as  $\bx-\bb$,
 to be predicted as the same class as the original label $t_0=\arg \max_i [\zeta(\bx_0)]_i$.% The loss function is defined as:
%\begin{align}
%\label{eqn_loss_f_pos}
%&f^{\textnormal{pos}}_\kappa(\bx_0,\bdelta)\nonumber\\&= 
%\max \{  \max_{i \neq t_0} [\zeta(\bx_0+\bdelta)]_i - [\zeta(\bx_0+\bdelta)]_{t_0}, - \kappa   \}.
%\end{align}
The loss function %$f^{\textnormal{pos}}_\kappa$
is a hinge-like loss and
the term $\kappa \geq 0$  controls the gap between $[\zeta(\bx_0+\bdelta)]_{t_0}$ and the other most probable class. In particular, the loss attains its minimal value when $[\zeta(\bx_0+\bdelta)]_{t_0}$ is $\kappa$ larger than $\max_{i \neq t_0} [\zeta(\bx_0+\bdelta)]_i$. The parameter $c \geq 0$ is the regularization coefficient associated with the first term.%$f^{\textnormal{pos}}_\kappa$.
The second and third terms in \eqref{eqn_per_pos} are jointly called the elastic-net regularizer \cite{zou2005regularization}, which aids in selecting a set of highly relevant features from $\bx-\bb$, and the parameter $\beta \geq 0$ controls the sparsity of the vector $\bx-\bb$.% In other words, if the $i$-th element of $\bx-\bb$ is $0$, this means the $i$-th is not significant for constituting PP.

Optionally, the input distribution $p(\bx)$ also maybe estimated from the data %using standard density estimators or for high dimensional data using copulas,
which could be used to further direct the search so that we produce realistic or high probability $\bx$.

\subsection{Computing Pertinent Negatives}
Analogous to PP, for PN let $\Delta_{PN}$ denote the space $\{\bdelta:  | \bx_0+\bdelta - \bb | 	\succ   | \bx_0 - \bb |  \textnormal{~and~} \bx_0 + \bdelta \in \cX    \}$. To solve for PN, we propose the following problem formulation:
\begin{align}
\label{eqn_per_neg}
\bdelta^{\textnormal{neg}}\leftarrow\argmin_{\bdelta \in \Delta_{PN}}  ~c \cdot\max \{ [\zeta(\bx_0+\bdelta)]_{t_0} - \max_{i \neq t_0} [\zeta(\bx_0+\bdelta)]_i, - \kappa   \}+\beta \|\bdelta  \|_1 + \|\bdelta \|_2^2- \gamma p(\bx_0 + \bdelta)
\end{align}
%\begin{align}
%\label{eqn_loss_f_neg}
%&\text{where, } f^{\textnormal{neg}}_\kappa(\bx_0,\bdelta)\nonumber\\&= \max \{ [\zeta(\bx_0+\bdelta)]_{t_0} - \max_{i \neq t_0} [\zeta(\bx_0+\bdelta)]_i, - \kappa   \}.
%\end{align}
In other words, for PN, we aim to find the least modified changes in $\bdelta \in \Delta_{PN}$, evaluated by the elastic-net loss on $\bdelta$,
such that its addition to $\bx_0$ leads to a different prediction from $t_0$, 

\subsection{Method Details}

We now describe the details of how the optimization of the above objectives is implemented along with estimation and modeling of certain key aspects.%\footnote{Code part of product, so unfortunately, couldn't be provided.}.

\subsubsection{Optimization Procedure}

 %If the gradient of the designed loss functions for PP and PN with respect to $\bdelta$ can be obtained, then one can readily apply a projected fast iterative shrinkage-thresholding algorithm (FISTA) \cite{beck2009fast} to solve problems (\ref{eqn_per_neg}) and (\ref{eqn_per_pos}) efficiently. However, in our black-box setting such gradient is inadmissible. We will illustrate how to get around this problem in the following subsection. 

Here we first illustrate how FISTA solves for PP and PN, assuming the gradient is available. This is very similar to previous work \cite{CEM} with the main difference lying in the projection operators $\Delta_{PN}$ and $\Delta_{PP}$.
 FISTA is an efficient solver for optimization problems involving $L_1$ regularization. Take pertinent negative as an example, let $g (\bdelta) = \max \{ [\zeta(\bx_0+\bdelta)]_{t_0} - \max_{i \neq t_0} [\zeta(\bx_0+\bdelta)]_i, - \kappa   \} + \|\bdelta\|_2^2 -  \gamma p(\bx_0 + \bdelta) $ denote the objective function of (\ref{eqn_per_neg}) without the $L_1$ regularization term.
Given the initial iterate $\bdelta^{(0)}=\mathbf{0}$,
projected FISTA iteratively updates the perturbation $I$ times by
\begin{equation}
\bdelta^{(k+1)}=\Pi_{\Delta_{PN}} \{ S_{\beta}(\by^{(k)}-\alpha_k \nabla g(\by^{(k)})) \};~~~
 \by^{(k+1)}=\Pi_{\Delta_{PN}} \{ \bdelta^{(k+1)}+\frac{k}{k+3} (\bdelta^{(k+1)} - \bdelta^{(k)}) \},
\end{equation}
where $\Pi_{\Delta_{PN}}$ denotes the vector projection onto the set $\Delta_{PN}$,
$\alpha_k$ is the step size, $\by^{(k)}$ is a slack variable accounting for momentum acceleration with $\by^{(0)}=\bdelta^{(0)}$, and $S_{\beta}: \bbR^{p} \mapsto \bbR^{p}$ is an element-wise shrinkage-thresholding function which is 0 if $\forall i \in \{1,\ldots,d\}$ $|\bz_i| \leq \beta$, else takes the values $\bz_i - \beta$ or $\bz_i + \beta$ for $\bz_i  > \beta$ or $\bz_i  < -\beta$ respectively.
%\begin{align}
%\label{eqn_ISTA_S}
%[S_{\beta}(\bz)]_i= \left\{
%\begin{array}{ll}
%\bz_i - \beta, & \text{~if~}\bz_i  > \beta ; \\
%0, & \text{~if~} |\bz_i| \leq \beta ; \\
%\bz_i + \beta, & \text{~if~}\bz_i  < -\beta,
%\end{array}
%\right.  
%\end{align}
%for any $i \in \{1,\ldots,d\}$.
The final perturbation $\bdelta^{(k^*)}$ for pertinent negative analysis is selected from the set $\{ \bdelta^{(k)}\}_{k=1}^I$ such that $f^{\textnormal{neg}}_{\kappa}(\bx_0,\bdelta^{(k^*)})=0$ and  $k^* = \arg \min_{k \in \{1,\ldots, I\}} \beta \| \bdelta\|_1 + \| \bdelta\|_2^2 $.
A similar projected FISTA approach is applied to PP analysis.

\subsubsection{Gradient Estimation}
In the black-box setting, in order to balance the model query complexity and algorithmic convergence rate using zeroth-order optimization, in this paper we use a two-point evaluation based gradient estimator averaged over $q$ different random directions \cite{duchi2015optimal,liu2017zeroth,liu2018zeroth}. Specifically, given a scalar function $f(\cdot)$, its gradient at a point $\bx \in \bbR^d$ is estimated by 
\begin{align}
\label{eqn_grad_est}
\widehat{\nabla} f (\bx) = \frac{d}{q \mu} \sum_{j=1}^q \frac{f(\bx+\mu \bu_j) - f(\bx)}{\mu} \cdot \bu_j,
\end{align}
where $\{\bu_j\}_{j=1}^q$ is a set of i.i.d. random directions drawn uniformly from a unit sphere, and $\mu>0$ is a smoothing parameter.

The estimation error between $\widehat{\nabla} f (\bx)$ and the true gradient $\nabla f(\bx)$ can be analyzed through a smoothed function $f_\mu (\bx)=\bbE_{\bu \in U_b} [f(\bx+\mu \bu)]$, where $U_b$ is a uniform distribution over the unit Euclidean ball. Assume $f$ is an $L$-smooth function, that is, its gradient $\nabla f$ is $L$-Lipschitz continuous.
It has been shown in \cite{liu2018zeroth} that $\widehat{\nabla} f (\bx)$ is an unbiased estimator of the gradient $\nabla f_\mu (\bx)$, i.e., $\bbE_{\bu} [\widehat{\nabla} f (\bx)]=\nabla f_{\mu} (\bx)$. Moreover, using the bounded error between $f$ and $f_\mu$, one can show that  the mean squared error between $\widehat{\nabla} f (\bx)$ and  $\nabla f (\bx)$ is upper bounded by 
\begin{align}
\label{eqn_bound}
\bbE_{\bu} [ \| \widehat{\nabla} f (\bx)- \nabla f (\bx)\|_2^2 ]\leq O \left(\frac{q+d}{q} \right) \|\nabla f (\bx)\|_2^2 + O(\mu^2 L^2 d^2).
\end{align}

%how gradient is estimated along with theoretical result about bounded bias (and variance) ...

\subsubsection{Determining Base Values}
As mentioned before, ideally, we would want base values as well as allowed ranges or limits for all features be specified by the user. This should in all likelihood provide the most useful explanations. However, this may not always be feasible given the dimensionality of the data and the level of expertise of the user. In such situations we compute base values using our best judgment.

For real valued features, we set the base value to be the median value of the feature. This possibly is the least interesting value for that feature as well as being robust to outliers.% Moreover, medians are known to be robust to outliers and are thus preferable to using means. They also are a point estimate that has minimum $L_1$ error w.r.t. the values for that feature.  
Medians also make intuitive sense where for sparse features $0$ would rightly be chosen as the base value as opposed to some other value which would be the case for means.
\begin{wrapfigure}{r}{0.5\textwidth}
  \centering  
      \includegraphics[width=0.5\textwidth]{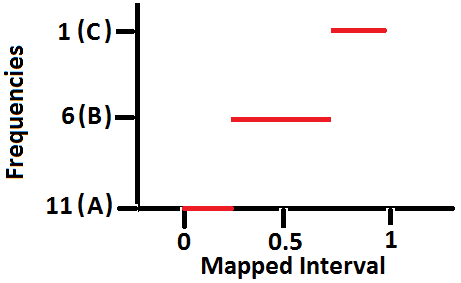}  
  \caption{Above we see a categorical feature taking three values A, B and C with frequencies 11, 6 and 1 respectively as indicated on the vertical axis. Our mapping function in equation 11 for FMA maps these frequencies and hence the categorical values to 0, 0.5 and 1 in the $[0,1]$ interval. The red horizontal lines depict the function $h(.)$ showcasing the range of values that map back to either A, B or C.}
  \label{FMA}
\vspace{-0.75cm}
\end{wrapfigure}
For categorical features, we set the base value to be the mode for that feature. Here we use the fact that the mode is the least informative value for a categorical feature \cite{inftheory} and rarer values are likely to be more interesting to the user. For example in a dataset containing health records most people will probably not have cancer and so having cancer is something that should stand out as it indicates a state away from the norm. Such states or behaviors we believe carry information that is more likely to surprise the user and draw attention, which could be a prelude to further actions.

\subsubsection{Modeling Categorical Features}
Given that categorical features do not impose an explicit ordering like real features do along with the fact that only the observed values have semantic meaning, there is a need to be model them differently when obtaining explanations. We now present a strategy that accomplishes this in an efficient manner compared to one-hot encoding where one could have an explosion in the feature space if there are many categorical features with each having many possible discrete values. Both these strategies are described next.%The alternate one-hot encoding approach is described in the supplement along with a discussion of the tradeoffs.

%two different strategies for handling categorical features in our framework along with a formal interpretation of what these strategies are actually doing.\\

%\begin{figure}[t]
 % \centering  
  %    \includegraphics[width=0.6\textwidth]{FMA.png}  
  %\caption{Above we see a categorical feature taking three values A, B and C with frequencies 11, 6 and 1 respectively as indicated on the vertical axis. Our mapping function in equation \ref{stgy1} for FMA maps these frequencies and hence the categorical values to 0, 0.5 and 1 in the $[0,1]$ interval. The red horizontal lines depict the function $h(.)$ showcasing the range of values that map back to either A, B or C.}
  %\label{FMA}
%\vspace{-0.5cm}
%\end{figure}

\noindent{\textbf{Frequency Map Approach (FMA):}} In this approach we want to directly leverage the optimization procedure described above where we need to define an ordered set/interval in which to find the perturbations $\delta$ for both PPs and PNs.

\noindent\textit{Mapping:} As described above for categorical features we set the base value to be the mode of the values that occur. Given this a natural ordering can be created based on the frequencies of the different values. Thus, the least frequent value would be considered to be the farthest from the base value. Based on this we can map the $k$ discrete values $v_{i1}, ..., v_{ik}$  of the $i^{th}$ feature occurring with frequencies/counts $c_{i1}, ..., c_{ik}$ to real values ${r}_{i1}, ..., {r}_{ik}$ respectively in the $[0,1]$ interval using the following mapping for any $j\in \{1, ..., k\}$:
%\begin{equation}
%\label{stgy1}
   $ {r}_{ij}=\frac{c_{\text{max}}-c_{ij}}{c_{\text{max}}-1}$
%\end{equation}
where $c_{\text{max}}=\max\limits_{j\in \{1, ..., k\}} c_{ij}$. This maps the discrete value with the highest frequency to 0 making it the base value, while all other values with decreasing frequencies lie monotonically away from 0. Every candidate value has to have a frequency of at least 1 and so every discrete value gets mapped to the $[0,1]$ interval. We divide by $c_{\text{max}}-1$, rather than $c_{\text{max}}-c_{\text{min}}$, where $c_{\text{min}}=\min\limits_{j\in \{1, ..., k\}} c_{ij}$ since, we do not want values that occur with almost equal frequency to be pushed to the edges of the interval as based on our modeling they are of similar interest. %For example, a feature may just have two values occurring with frequencies of 50 and 49. We want to be able to switch between these values easily when finding PPs and PNs rather than mapping them to two ends of the interval i.e. 50 to 0 and 49 to 1. This is particularly important when we notice the fact that a different categorical feature may have more extreme frequencies such as 50 and 10, and we want to maintain this relative difference between features in our mapping.

\noindent\textit{Method and Interpretation:} Based on the equation for $r_{ij}$, we run our algorithm for categorical features in the interval $[0,1]$, where every time we query the model we round the $\bx_0+\bdelta$ to the closest $r_{ij}$ so that a \emph{valid} categorical value $v_{ij}$ can be mapped back to and sent as part of the query input.

The question now is what are we exactly doing in the mathematical sense. It turns out that rather than optimizing $f^{\textnormal{neg}}_{\kappa}(\bx_0,\bdelta)=\max \{ [\zeta(\bx_0+\bdelta)]_{t_0} - \max_{i \neq t_0} [\zeta(\bx_0+\bdelta)]_i, - \kappa   \}$ or $f^{\textnormal{pos}}_{\kappa}(\bx_0,\bdelta)=\max \{  \max_{i \neq t_0} [\zeta(\bx_0+\bdelta)]_i - [\zeta(\bx_0+\bdelta)]_{t_0}, - \kappa   \}$, we are optimizing $f^{\textnormal{neg}}_{\kappa}(h(\bx_0,\bdelta))$ or $f^{\textnormal{pos}}_{\kappa}(h(\bx_0,\bdelta))$ respectively, where $h(.)$ is the identity map for real features,
% that is $f^{\textnormal{neg}}_{\kappa}(h(\bx_0,\bdelta))=f^{\textnormal{neg}}_{\kappa}(\bx_0,\bdelta$) and $f^{\textnormal{pos}}_{\kappa}(h(\bx_0,\bdelta))=f^{\textnormal{pos}}_{\kappa}(\bx_0,\bdelta$),
but a step function defined over the $[0,1]$ interval for categorical features. Let $h_i(.)$ denote the application of the function $h(.)$ to the categorical feature $i$. If $\bx = \bx_0+\bdelta$ and $\bx_i$ denotes the value of the feature in the mapped $[0,1]$ interval then,
%\begin{align}
%\label{gstep}
    $h_i(\bx_0,\bdelta) = v_{ij}$, if $|\bx_i-r_{ij}| \le |\bx_i-r_{im}|$ $\forall m\in\{1, ..., k\}$ and $m\neq j$
%\end{align}
where $|.|$ denotes absolute value.
An example function $h(.)$ is depicted in figure \ref{FMA}, where we see how real values are mapped back to valid categorical values by rounding to the closest $r_{ij}$.

\noindent\textbf{Simplex Sampling Approach (SSA):} In this method of handling categorical variables, we will assume that a one-hot encoding of the input. Let $\mathbf{x}=[\mathbf{x}_C~\mathbf{x}_R]$ be the input feature vector where $\mathbf{x}_C$ denotes the categorical part while $\mathbf{x}_R$ denotes the set of real features. Let there be $C$ categorical features in total. Then for all $c \in [1:C], x_c \in [1:d_c]$ where $x_c$ is the $c$-th categorical variable and it takes one of $d_c$ values.
 Note that we imply no ordering amongst the $d_c$ values. Generically they are assumed to have distinct integer values from $1$ till $d_c$. 
We assume that input $\mathbf{x}$ is processed into $\mathbf{\tilde{x}} = [\mathbf{\tilde{x}}_C ~ \mathbf{\tilde{x}}_R] $ where $\mathbf{\tilde{x}}_R = \mathbf{x}_R$ while $\mathbf{\tilde{x}}_C \in \mathbb{R}^{1 \times \prod_{c \in C} d_c }$. Each component $\mathbf{\tilde{x}}_c \in \mathbb{R}^{1 \times d_c} $ is set to $\mathbf{e}_i$, the canonical unit vector with $1$ in the $i$-th coordinate, if $x_c$ takes the value $i$. 

\textit{Interpretation:} Now, we provide an interpretation when every categorical component $c$ lies in the $d_c$ dimensional simplex, i.e. when $\mathbf{\tilde{x}}_c \in \Delta_{d_c} $. Here, $\Delta_N$ denotes the $N$-dimensional simplex. The actual function can be evaluated only on the inputs where each categorical component takes values from one of the corner points on the simplex, namely $\mathbf{e}_i, ~ i \in [1:d_c]$. Therefore, we interpolate the function when $\mathbf{\tilde{x}}_c$ is assigned a real vector in the simplex. 

Let $f(\cdot)$ capture the soft output of the classifier when the one-hot encoded categorical variables take the values at the corner points of their respective simplices. Now, we extend the definition of $f$ as follows:

$f([\mathbf{\tilde{x}}_C~\mathbf{x}_R]) = 
\mathbb{E}_{\mathbf{e}_{i_c} \sim \mathbf{\tilde{x}}_c,~\forall c \in [1:C] } \left[ f(\mathbf{e}_{i_1}, \ldots \mathbf{e}_{i_c}, \ldots \mathbf{e}_{i_{C}}, \mathbf{x}_R)\right].$

Essentially, we sample the $c$-th unit vector from the distribution represented by $\mathbf{\tilde{x}_c} \in \Delta_{d_c}$ on the simplex independent of other categorical variables. The function value is the expected value of the functional evaluated on unit vectors obtained from this product distribution along with the fixed real coordinates $\mathbf{x}_R$.

When we perform the gradient descent as a part of algorithm \ref{cem}, we actually do a projected gradient descent for $\mathbf{\tilde{x}}_C $ in the product of simplices $\Delta_{d_1} \times \ldots \Delta_{d_c}$. We cannot evaluate the function exactly, hence we can average over a certain number of samples drawn from the product distribution for every function evaluation on a candidate $\mathbf{\tilde{x}}$.\\

\noindent\textbf{FMA vs SSA Tradeoffs:} As such the SSA strategy has stronger theoretical grounding, however from a practical standpoint it requires a lot of additional averaging through sampling for every function evaluation along with repeated projections to the simplices defined for every categorical feature during gradient descent to optimize the objective in Algorithm \ref{cem}. 
FMA can also take more general format of inputs as they don't need to be one-hot-encoded which guards against further explosion of the feature space which could potentially result from categorical features having many distinct values.

%\vspace{-0.3cm}
\section{Experiments}
%\vspace{-0.3cm}
We now empirically evaluate our approach on 5 public datasets covering diverse domains. The datasets along with their domains are as follows. Finance: German Credit \cite{uci} and FICO \cite{FICO}, Astronomy: Digital Sky Survey \cite{kaggleskysurvey}, Health care: Vertebral Column \cite{uci} and Neuroscience: Olfaction \cite{olfs}. A summary of the datasets is given in table \ref{tab:data}. All the datasets except olfaction have class labels. However, as done in previous studies \cite{tip}, we use \emph{Pleasantness} as the target which is binarized using a threshold of 50. That is molecules with a rating $>$ 50 (on a 100 point scale) are categorized as being pleasant to smell, while the rest are deemed as unpleasant.

\begin{wraptable}{r}{0.6\textwidth}
\vspace{-0.5cm}
\centering
\caption{Dataset characteristics, where $N$ denotes dataset size and $d$ is the dimensionality.}
 \begin{tabular}{|c|c|c|c|c|}
  \hline
    \small{Dataset} & \small{$N$} & \small{$d$} & \small{\# of} & \small{Domain} \\
    &&  & \small{Classes}& \\
    \hline
     \small{German Credit} & \small{1000} & \small{20} & \small{2} & \small{Finance} \\
     \hline
    \small{FICO} & \small{10459} & \small{24} & \small{2} & \small{Finance} \\
 \hline
   \small{Sky Survey} & \small{10000} & \small{17} & \small{3} & \small{Astronomy} \\
   \hline
 \small{Vertibral Column} & \small{310} & \small{6} & \small{3} & \small{Health care} \\
 \hline
\small{Olfaction} & \small{476} & \small{4869} & \small{2} & \small{Neuroscience} \\
 \hline
 \end{tabular}
 \label{tab:data}
\vspace{-0.25cm}
\end{wraptable}
We test the methods for two (black box) classification models namely, CART decision trees (depth $\le$ 5) and random forest (size 100). In all cases we use a random 75\% of the dataset for training and a remaining 25\% as test. We repeat this 10 times and average the results. The explanations are generated for the test points and correspondingly evaluated. Given the generality and efficiency of the FMA approach as stated above, we use that in our implementation of MACEM. For all datasets and all features the ranges were set based on the maximum and minimum values seen in the datasets. The base values for the German Credit, Sky Survey, Vertibral  Column dataset and FICO were set to median for the real features and mode for the categorical ones. For FICO the special values (viz. -9, -7, -8) were made 0. The base values for all features in the olfaction dataset were set at 0. We do not learn the underlying distribution so $\gamma$ is set to 0 and the other parameters are found using cross-validation. We generate 50 random samples for estimating gradients at each step and run our search for 100 steps. We compare with LIME (\url{https://github.com/marcotcr/lime}) which is arguably the most popular method to generate local explanations in a model agnostic fashion for (especially non-neural network) classifiers trained on structured data. We quantitatively evaluate our results on all 5 datasets as described next. We also provide qualitative evaluations for German Credit and the Olfaction dataset based on studying specific cases along with obtaining expert feedback.
%\begin{table}[t]
%\centering
%\caption{Dataset characteristics, where $N$ denotes dataset size and $d$ is the dimensionality.}
 %\begin{tabular}{|c|c|c|c|c|}
  %\hline
   % \small{Dataset} & \small{$N$} & \small{$d$} & \small{\# of} & \small{Domain} \\
    %&&  & \small{Classes}& \\
    %\hline
     %\small{German Credit} & \small{1000} & \small{20} & \small{2} & \small{Finance} \\
     %\hline
    %\small{FICO} & \small{10459} & \small{24} & \small{2} & \small{Finance} \\
 %\hline
  % \small{Sky Survey} & \small{10000} & \small{17} & \small{3} & \small{Astronomy} \\
   %\hline
 %\small{Vertibral Column} & \small{310} & \small{6} & \small{3} & \small{Health care} \\
 %\hline
%\small{Olfaction} & \small{476} & \small{4869} & \small{2} & \small{Neuroscience} \\
 %\hline
 %\end{tabular}
 %\label{tab:data}
%\vspace{-0.5cm}
%\end{table}
\vspace{-0.2mm}
\subsection{Quantitative Evaluation Metrics}
\vspace{-0.2mm}
We define three quantitative metrics: Correct Classification Percentage (CCP), Correct Feature Ranking (CFR) and Correct Feature Importance Percentage (CFIP) where we evaluate feature importances as well as how accurate the explanations are in predicting the same or different classes for PPs and PNs respectively. %The description as well as experimental results for the CFIP are given in the supplement.

For LIME we create proxies for PPs and PNs that are intuitively similar to ours for a fair comparison. The PP proxy for LIME is created by replacing all the negatively correlated features by base values in the original example, while maintaining the positively correlated feature values. For PNs, we create a proxy by setting all the positively correlated feature values to base values, while maintaining the negatively correlated features. % This is intuitively less similar to our PNs as the unchanged features in our case take the values of the original input, nonetheless it should mimic their behavior by lying in a different class.
%For our PPs and PNs we compute the feature importances in standard ways.
For our PPs, we compute feature importance by taking the absolute difference of each feature value to the corresponding base value and dividing by the features standard deviation. For PNs, we take the absolute value of the change of the perturbed feature and divide again by standard deviation. For LIME, we use the absolute value of the coefficients of the PP/PN proxies.

\noindent\textbf{Correct Classification Percentage (CCP):} For this metric we compute what percentage of our PPs actually lie in the same (predicted) class as the original example and for PNs what percentage of the new examples lie in a different class than the predicted one of the original example. Formally, if $(x_1,t_1), ..., (x_n,t_n)$ denote $n$ examples with $t_i$ being the predicted class label for $x_i$ and $PP_i$, $PN_i$ being the respective pertinent positives and negatives for it, then if $\lambda(.)$ denotes an indicator function which is 1 if the condition inside it is true and 0 otherwise, we have (higher values better)
\begin{align}
\label{ccp}
    CCP_{PP}=\sum_i\frac{\lambda\left(\max [\zeta(PP_i)]=t_i\right)}{n}\times 100, ~~~
    CCP_{PN}=\sum_i\frac{\lambda\left(\max [\zeta(PN_i)]\neq t_i\right)}{n}\times 100
\end{align}
%Higher values for both $CCP_{PP}$ and $CCP_{PN}$ are desirable.

\noindent\textbf{Correct Feature Ranking (CFR):} For this metric we want to evaluate how good a particular explanation method's feature ranking is. For PPs/PNs we independently set the top-k features to base/original values and in each case note the class probability of classifying that input into the black box models predicted class. We then rank the features based on these probabilities in descending order and denote this ranking by $r^*_{PP}$ (or $r^*_{PN}$). Our CFR metric is then a correlation between the explanation models ranking of the features ($r_{PP}$ or $r_{PN}$) and this ranking. Higher the CFR better the method. If $\rho(.,.)$ indicates the correlation between two lists then,
\begin{equation}
CFR_{PP} = \rho(r_{PP},r^*_{PP}),~~~CFR_{PN} = \rho(r_{PN},r^*_{PN})
\end{equation}
The intuition is that most important features if eliminated should produce the most drop in predicting the input in the desired class. This is similar in spirit to ablation studies for images \cite{Ormas} or the faithfulness metric proposed in \cite{selfEx}.

\noindent\textbf{Correct Feature Importance Percentage (CFIP):} Here we want to validate if the features that are actually important for an ideal PP and PN are the ones we identify. %Comparing with the features in the tree path of the original input we want to explain is not the ideal comparison as we are creating new inputs with minimal variation that either maintain or change class. Hence, although the features that arise as important for PPs and PNs appear in the original inputs path all of them are not required to be present there. A better comparison is with feature sets that would match the ideal PPs and PNs for that input. Saying this though, we also provide examples of the tree path for some inputs and the corresponding PPs and PNs, showcasing how the important features identified for these PPs and PNs correspond to features in the tree paths at different levels.
Of course, we do not know the globally optimal PP or PN for each input. So we use proxies of the ideal to compare with, which are training inputs that are closest to the base values and satisfy the PP/PN criteria as defined before. % For PPs the proxy we use for the ideal PPs are training inputs that are closest to the base values but lie in the same class as the original input. For PNs we find training inputs that lie in a different class than the original inputs but which are closest to the original inputs, while being farther away from the base values relative to these inputs. %In other words, we are finding training inputs that are slightly farther away than the original inputs but lie in a different class. 
If our (correct) PPs and PNs are closer to the base values than these proxies, then we use them as the golden standard. We compute the CFIP score as follows: Let $f^*_{PP}(x)$ and $f^*_{PN}(x)$ denote the set of features in the tree corresponding to the ideal (proxy) PPs and PNs for an input $\bx$ as described before. Let $f_{PP}(x)$ and $f_{PN}(x)$ denote the top-k important features (assuming $k$ is tree path length) based on our method or LIME then,
\begin{align}
\label{cfi}
    CFIP_{PP}=\frac{100}{n}\sum_i\frac{|f_{PP}(x_i)\cap f^*_{PP}(x_i)|_{\text{card}}}{k}, ~~~
    CFIP_{PN}=\frac{100}{n}\sum_i\frac{|f_{PN}(x_i)\cap f^*_{PN}(x_i)|_{\text{card}}}{k}
\end{align}
where, $|.|_{\text{card}}$ denotes cardinality of the set. Here too higher values for both $CFIP_{PP}$ and $CFIP_{PN}$ are desirable.
\vspace{-0.3cm}
\begin{table*}[htbp]
\centering
\caption{Below we see the quantitative results for CCP metric. The statistically significant best results are presented in bold based on paired t-test.}
 \begin{tabular}{|c|c|c|c|c|c|c|c|c|}
  \hline
    \multirow{2}{*}{\small{Dataset}} & \multicolumn{2}{c|}{\small{$CCP_{PP}$ (Tree)}} & \multicolumn{2}{c|}{\small{$CCP_{PN}$ (Tree)}} & \multicolumn{2}{c|}{\small{$CCP_{PP}$ (Forest)}} & \multicolumn{2}{c|}{\small{$CCP_{PN}$ (Forest)}} \\
     & \small{MACEM} & \small{LIME} & \small{MACEM} & \small{LIME} & \small{MACEM} & \small{LIME}  & \small{MACEM} & \small{LIME}\\
     \hline
     \small{German Credit} & \small{\textbf{100}} & \small{96.0} & \small{\textbf{100}} & \small{10.2} & \small{\textbf{100}} & \small{89.6} & \small{\textbf{100}} & \small{9.0} \\
 \hline
    \small{FICO} & \small{\textbf{100}} & \small{91.45} & \small{\textbf{100}}& \small{49.31} & \small{\textbf{100}} & \small{46.47} & \small{\textbf{100}} & \small{40.82}\\
 \hline
  \small{Sky Survey}& \small{\textbf{100}} & \small{58.48}& \small{\textbf{100}}& \small{25.01} & \small{\textbf{100}}& \small{47.32} & \small{\textbf{100}} & \small{39.78}\\
  \hline
 \small{Vertibral Column} & \small{\textbf{100}} & \small{33.33} & \small{\textbf{100}}& \small{44.87} & \small{\textbf{100}} & \small{91.02}& \small{\textbf{100}}& \small{19.23}\\
  \hline
\small{Olfaction} & \small{\textbf{100}} & \small{79.22} & \small{\textbf{100}}& \small{13.22} & \small{\textbf{100}} & \small{74.19}& \small{\textbf{100}}& \small{19.21}\\
 \hline
 \end{tabular}
 \label{tab:quant}
\vspace{-0.1cm}
\end{table*}
\vspace{-0.3cm}
\begin{table*}[htbp]
\centering
\caption{Below we see the quantitative results for CFR metric. The statistically significant best results are presented in bold based on paired t-test.}
 \begin{tabular}{|c|c|c|c|c|c|c|c|c|}
  \hline
    \multirow{2}{*}{\small{Dataset}} & \multicolumn{2}{c|}{\small{$CFR_{PP}$ (Tree)}} & \multicolumn{2}{c|}{\small{$CFR_{PN}$ (Tree)}} & \multicolumn{2}{c|}{\small{$CFR_{PP}$ (Forest)}} & \multicolumn{2}{c|}{\small{$CFR_{PN}$ (Forest)}} \\
     & \small{MACEM} & \small{LIME} & \small{MACEM} & \small{LIME} & \small{MACEM} & \small{LIME}  & \small{MACEM} & \small{LIME}\\
     \hline
     \small{German Credit} & \small{\textbf{0.68}} & \small{0.64} & \small{\textbf{0.70}} & \small{0.68} & \small{\textbf{0.43}} & \small{0.35} & \small{\textbf{0.48}} & \small{\textbf{0.46}} \\
 \hline
    \small{FICO} & \small{\textbf{0.68}} & \small{0.49} & \small{\textbf{0.74}}& \small{0.70} & \small{\textbf{0.29}} & \small{0.09} & \small{\textbf{0.54}} & \small{0.31}\\
 \hline
  \small{Sky Survey}& \small{\textbf{0.55}} & \small{0.48}& \small{\textbf{0.81}}& \small{0.53} & \small{\textbf{0.42}}& \small{0.35} & \small{\textbf{0.54}} & \small{0.36}\\
  \hline
 \small{Vertibral Column} & \small{\textbf{0.63}} & \small{\textbf{0.64}} & \small{\textbf{0.75}}& \small{0.66} & \small{\textbf{0.20}} & \small{-0.02}& \small{\textbf{0.33}}& \small{0.23}\\
  \hline
\small{Olfaction} & \small{\textbf{0.71}} & \small{0.58} & \small{\textbf{0.78}}& \small{0.59} & \small{\textbf{0.73}} & \small{0.62}& \small{\textbf{0.82}}& \small{0.65}\\
 \hline
 \end{tabular}
 \label{tab:quant2}
\vspace{-0.1cm}
\end{table*}

\begin{table*}[htbp]
\centering
\caption{Below we see the quantitative results for CFIP metric. The statistically significant best results are presented in bold based on paired t-test.}
 \begin{tabular}{|c|c|c|c|c|c|c|c|c|}
  \hline
    \multirow{2}{*}{\small{Dataset}} & \multicolumn{2}{c|}{\small{$CFIP_{PP}$ (Tree)}} & \multicolumn{2}{c|}{\small{$CFIP_{PN}$ (Tree)}} & \multicolumn{2}{c|}{\small{$CFIP_{PP}$ (Forest)}} & \multicolumn{2}{c|}{\small{$CFIP_{PN}$ (Forest)}} \\
     & \small{MACEM} & \small{LIME} & \small{MACEM} & \small{LIME} & \small{MACEM} & \small{LIME}  & \small{MACEM} & \small{LIME}\\
     \hline
     \small{German Credit} & \small{\textbf{83.73}} & \small{68.98} & \small{\textbf{62.32}} & \small{48.28} & \small{\textbf{71.26}} & \small{30.65} & \small{\textbf{30.13}} & \small{13.98} \\
 \hline
    \small{FICO} & \small{\textbf{79.13}} & \small{78.96} & \small{\textbf{82.92}}& \small{58.76} & \small{\textbf{89.92}} & \small{33.56} & \small{\textbf{54.98}} & \small{33.25}\\
 \hline
  \small{Sky Survey}& \small{\textbf{98.05}} & \small{76.55}& \small{\textbf{93.52}}& \small{68.55} & \small{\textbf{78.96}}& \small{\textbf{80.18}} & \small{\textbf{89.54}} & \small{\textbf{88.26}}\\
  \hline
 \small{Vertibral Column} & \small{\textbf{85.95}} & \small{68.24} & \small{\textbf{77.38}}& \small{64.98} & \small{\textbf{94.23}} & \small{86.12}& \small{\textbf{100.00}}& \small{89.45}\\
  \hline
\small{Olfaction} & \small{\textbf{87.19}} & \small{83.92} & \small{\textbf{72.82}}& \small{47.96} & \small{\textbf{82.23}} & \small{77.56}& \small{\textbf{74.72}}& \small{52.28}\\
 \hline
 \end{tabular}
 \label{tab:quant3}
\vspace{-0.1cm}
\end{table*}

%\subsection{Results}
%We now discuss how our method fairs quantitatively and for a couple of datasets also qualitatively.
\vspace{-0.2cm}
\subsection{Quantitative Evaluation}
\vspace{-0.1cm}
First looking at Table \ref{tab:quant} we observe that our method MACEM as designed, on all datasets produces PPs and PNs that lie in the same or different class as the original input respectively. This is indicated by metrics $CCP_{PP}$ and $CCP_{PN}$ where we are 100\% accurate. This observation is reassuring as it means that whenever we return a PP or PN for an input it is valid. Although LIME has reasonable performance for PPs (much worse for PNs) no such promise can be made.
\begin{wrapfigure}{r}{0.49\textwidth}
  \centering  
      \includegraphics[width=0.49\textwidth]{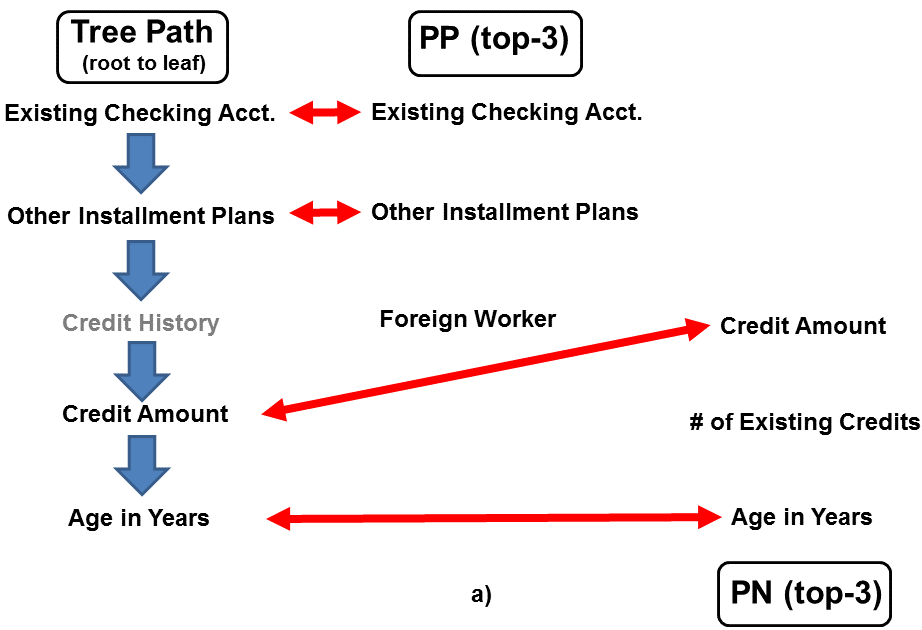}\\  
      \includegraphics[width=0.49\textwidth]{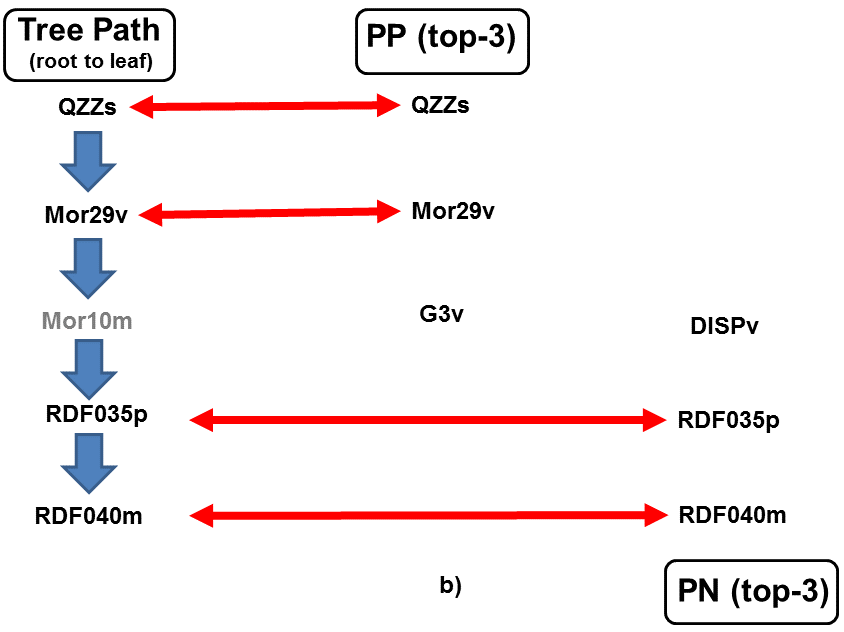}
      \vspace{-0.5cm}
  \caption{Above we compare the actual tree path (blue arrows) and the corresponding PP/PN important features for an input in the a) German Credit dataset and b) Olfaction dataset. The PP columns (center) and the PN columns (right) list the top 3 features highlighted by MACEM for the corresponding PPs and PNs. The PP feature importance reduces top to bottom, while the PN feature importance reduces bottom to top. The red arrows indicate PP and PN features that match the features in the tree path for the respective inputs. }
  \label{GCeg}
\vspace{-1cm}
\end{wrapfigure}
Looking at Table \ref{tab:quant2} we observe that the feature ranking obtained by our method seems to be more indicative of the true feature importances than LIME, as it correlates better with the ablation based ranking in almost all cases as conveyed by the higher CFR score. We are the best performer in most cases with high correlation values depicting that we accurately capture (relative) feature importances.

We now look at how efficient the different methods are in picking the correct features for PPs and PNs. This is reflected in the $CFIP_{PP}$ and $CFIP_{PN}$ metrics in Table \ref{tab:quant3}. We are still better than LIME in both cases. Our PP results are generally better than PN. A reason for this maybe that when we run our algorithm (based on FISTA) most initial movements, especially for inputs in the interior of the decision boundary, will result in not changing the class, so the candidate PNs to choose from eventually (where we output the sparsest) will be much smaller than the candidate PPs for many inputs. This leads to a much richer choice from which to choose the (sparsest) PP resulting in them being more accurate.% As such though, our PNs will still necessarily lie in a different class as seen before.

\vspace{-0.2cm}
\subsection{Qualitative Evaluation}
\vspace{-0.1cm}

We now look at specific inputs from two of the datasets and garner expert feedback relating to the insights conveyed by the models and our explanations for them.

The smallest change to alter the class will probably occur by changing values of the lower level features in the tree path, which is what we observe. Hence, both PPs and PNs seem to capture complementary information and together seem to be important in providing a holistic explanation.
\noindent\textbf{Comparing with Tree Paths:} In figure \ref{GCeg}, we see two examples the left from German Credit dataset and the right from Olfaction dataset. In both these examples we compare the top features highlighted by our method with actual tree paths for those inputs. In figure \ref{GCeg}a, we see that our top-2 PP features coincide with the top-2 features in the tree path. While the bottom-2 features in the tree path correspond to the most and $3^{rd}$ most important PN feature. A qualitatively similar result is seen in figure \ref{GCeg}b for the olfaction dataset. We observe this for over 80\% inputs in these datasets. What this suggests is that our PP features tend to capture more global or higher level features, while our PNs capture more detailed information. This makes intuitive sense since for PPs we are searching for the sparsest input that lies in a particular class and although the input (absolute) feature values have to upper bound the PPs feature values, this is a reasonably weak constraint leading to many inputs in a class having the same PPs, thus giving it a global feel. On the other hand, a PN is very much input dependent as we are searching for minimal (addition only) perturbations that will change the class. 
%\vspace{-0.05cm}

\noindent\textbf{Human Expert Evaluations:} %Although typically in explainability research human evaluation is critical \cite{lime,CEM}, in our setup, since the models we built (small decision trees) were interpretable, human evaluation is much less important. Nonetheless,
We asked an expert in Finance and another in Neuroscience to validate our decision trees (so that their intuition is consistent with the model) and explanations. Each expert was given 50 randomly chosen explanations from the test set and they were blinded to which method produced these explanations. They were then given the binary task of categorizing if each explanation was reasonable or not.

The expert in Finance considered the features selected by our decision tree to be predictive. He felt that 41 of our PPs made sense, while around 34 PPs from LIME were reasonable. Regarding PNs, he thought 39 of ours would really constitute a class change, while 15 from LIME had any such indication. An expert from Neuroscience who works on olfaction also considered our tree model to be reasonable as the features it selected from the Dragon input features \cite{olfs} had relations to molecular size, paclitaxel and citronellyl phenylacetate, which are indicative of pleasantness. In this case, 43 of our PPs and 34 of LIMEs were reasonable. Again there was a big gap in PN quality, where he considered 40 of our PNs to be reasonable versus 18 of LIMEs. For random forests the numbers were qualitatively similar. For the finance dataset, 44 of ours and 27 of LIMEs PPs made sense to the expert. While 38 of our PNs and 19 of LIMEs were reasonable. For the olfaction dataset, 41 of our PPs and 32 of LIMEs were reasonable, while 39 of our PNs and 20 of LIMEs made sense.

\vspace{-0.4cm}
\section{Discussion}
\vspace{-0.2cm}
In this paper we provided a model agnostic black box contrastive explanation method specifically tailored for structured data that is able to handle real as well as categorical features in a meaningful and scalable manner. We saw that our method is quantitatively as well as qualitatively superior to LIME and provides more complete explanations.

In the future, we would like to extend our approach here to be applicable to also unstructured data. In a certain sense, the current approach could be applied to such data if it is already vectorized or the text is embedded in a feature space. In such cases, although minimally changing a sentence to another lying in a different class would be hard, one could identify important words or phrases which a language model or human could use as the basis for creating a valid sentence.% In the text domain one could also envision using sampling approaches developed in the multi-armed bandit literature to produce meaningful explanations as an alternative to zeroth order optimization which we have employed here.

\bibliography{ExAbsent}
\bibliographystyle{plain}

\end{document}